\documentclass[journal]{IEEEtran}
\ifCLASSINFOpdf
\else
\fi
\usepackage{mathtools}
\usepackage{color}
\usepackage{algorithm}
\usepackage[noend]{algpseudocode}

\usepackage{times}
\usepackage{epsfig}
\usepackage{graphicx}
\usepackage{amssymb}
\usepackage{romannum}
\usepackage{color}

\usepackage{amsmath}
\usepackage{breqn}
\usepackage{booktabs}

\begin{document}
%
\title{Systematic Comparative Analysis of Large Pretrained Language Models on Contextualized Medication Event Extraction}
%
%
%

 \author{Tariq Abdul-Quddoos, Xishuang Dong,  and~Lijun Qian,~\IEEEmembership{Senior Member,~IEEE}
\thanks{X. Dong, T.  Abdul-Quddoos, and L. Qian are with the Center of Excellence in Research and Education for Big Military Data Intelligence (CREDIT Center), Department of Electrical and Computer Engineering, Prairie View A\&M University, Texas A\&M University System, Prairie View, TX 77446, USA. Email: xidong@pvamu.edu, tabdulquddoos@pvamu.edu, liqian@pvamu.edu}
}

\maketitle

\begin{abstract}

Attention-based models have become the leading approach in modeling medical language for Natural Language Processing (NLP) in clinical notes. These models outperform traditional techniques by effectively capturing contextual representations of language.

In this research a comparative analysis is done amongst pre-trained attention based models namely Bert Base, BioBert, two variations of Bio+Clinical Bert, RoBerta, and Clinical Longformer on task related to Electronic Health Record (EHR) information extraction. The tasks from Track 1 of Harvard Medical School’s 2022 National Clinical NLP Challenges (n2c2) are considered for this comparison, with the Contextualized Medication Event Dataset (CMED) given for these task. CMED is a dataset of unstructured EHR's and annotated notes that contain task relevant information about the EHR's. The goal of the challenge is to develop effective solutions for extracting contextual information related to patient medication events from EHR's using data driven methods. 

 Each pre-trained model is fine-tuned and applied on CMED to perform medication extraction, medical event detection, and multi-dimensional medication event context classification. Processing methods are also detailed for breaking down EHR's for compatibility with the applied models. Performance analysis has been carried out using a script based on constructing medical terms from the evaluation portion of CMED with metrics including recall, precision, and F1-Score. The results demonstrate that models pre-trained on clinical data are more effective in detecting medication and medication events, but Bert Base, pre-trained on general domain data showed to be the most effective for classifying the context of events related to medications. 
 
 \end{abstract}


%
\IEEEpeerreviewmaketitle

\section{Introduction}
\label{sec1}
 An Electronic Health Record (EHR) holds records of patient health information generated by encounters in any care delivery setting, including information on patient demographics, progress notes, problem lists, medications, vital signs, past medical history, immunizations, laboratory data, and radiology reports~\cite{EHRBACKGROUND}. The application of machine learning on EHR's serves to place the burden of extracting information on data driven models rather than people. The National Clinical NLP Challenges (N2C2), hosted by Harvard Medical School, focuses on the study of applying data driven approaches to mining clinical information holding a number of challenges since 2006. Track 1 of the 2022 N2C2 has task researchers with developing solutions for extracting information related to the context of medication mentions in EHR's. For this challenge the Contextualized Medication Event Dataset(CMED) has been released and is a dataset capturing relevant context needed to understand medication changes in clinical narratives~\cite{N2C22022}, containing EHR's and annotated notes with task relevant information. In the competition, a number of methods were used with large language transformer based architectures, with some popular choices being Bert, RoBerta, DeBerta, Longformer, and Text-To-Text Transfer Transformer. In combination with these models additional architecture add-on methods consisted of multiple linear layers with combined losses and knowledge graphs~\cite{N2C22022RES}. 

In this work several pre-trained attention based language models undergo comparative analysis with information extraction task on CMED. The specific applied models are Bert Base, BioBert~\cite{BIOBERT}, two variations of Bio+Clinical Bert~\cite{BIOCLINBERT}, RoBerta Base~\cite{roberta}, and Clinical LongFormer~\cite{ClinLong} with pre-training from general, biomedical, and clinical domain corpora. The models are fine-tuned on CMED and used for the three task listed in section ~\ref{sec:challenges}. This study examines the performance of the applied models on the CMED dataset along with the differences in performance with regard to architecture differences,  pre-training methods, and pre-training corpora. Pre- and post- processing methods are also of presented in this study given the wide range of syntactic structures used in EHR’s, it can be difficult to judge how best the process an entire record.
The contributions of this work are as follow:
1) A comparative analysis of pre-trained large langauge attention models performance of EHR related task; 2) Fine-tuning pre-trained models on CMED training data for  medication detection,medication event detection, and multi-dimensional medication event context classification; 3) Processing methods for breaking down EHR's into format compatible with applied models. 

The rest of this paper is organized as follows. In Section ~\ref{sec:challenges} the three modeling task are detailed, in section ~\ref{methods} the applied modeling achitecture's are detailed along with the differnces between applied pre-trained models, in section~\ref{Experiment} an overview of processing methods is given and results for experiments are analyzed, in section~\ref{Related} an overview of related works is given, and in section~\ref{Conclusion} this work is summarized and future work of interest is detailed. 

\section{2022 National Clinical NLP Challenges}
\label{sec:challenges}
Track 1 of the 2022 N2C2 has task researchers with developing solutions for generating information related to the context of medication mentions in EHR's using data driven methods. A description of each task is given below. 
\subsection{Task 1: Medication Detection}
\label{sec:Task1_intro}
Task 1 for the 2022 N2C2 has task researchers with identifying medication mentions in EHR's. This task has been well studied by NLP researchers with it also being a task in the 2018 N2C2~\cite{N2C22018Track1}. Knowing what medications and where they are in EHR's are an essential step for further medication related information extraction.

\subsection{Task 2: Medication Event Classification}
\label{sec:Task2_intro}
Task 2 for the 2022 N2C2 task researchers with classifying identified medication mentions from task 1 as having events associated with them. An event refers to any change that has to do with a particular medication within its context. The three classes for this task are given as disposition, no disposition, and undetermined. Disposition refers to an event occurring with the associated medication, no disposition refers to no event occurring for the associated medication, and undetermined refers to if annotators cannot determine if an event has occurred or not~\cite{N2C22022}. 

\subsection{Task 3: Multi-dimensional Medication Event Context Classification}
\label{sec:Task3_intro}
Task 3 for the 2022 N2C2 task researchers with classifying the context of medications that have been labeled with disposition in task 2. The context is classified across 5 dimensions, those dimensions being action, temporality, certainty, and actor, and negation~\cite{N2C22022}. Action refers to type of change being discussed and has the following 7 classes: Start, Stop, Increase, Decrease, Unique Dose, Other Change, and Unknown. Temporality refers to when the change occurred and has the following 4 classes: Past, Present, Future, Unknown. Certainty refers to if a change was implemented or just discussed and has the following 4 classes: Certain, Hypothetical, Conditional, and Unknown. Actor refers to who initiated the change and has the following and has the following 3 classes: Physician, Patient, Unknown. Negation refers to if the medication event is negated and has the following two classes: negated and not negated.

\section{Method}
\label{sec4}
\label{methods}
\begin{figure}[H]
	 \centering
    	 \includegraphics[width=1\linewidth]{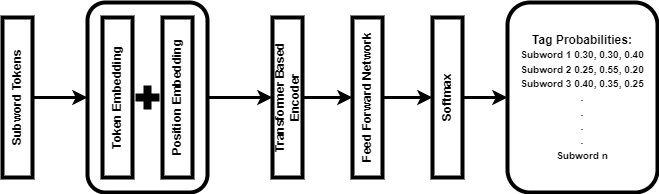}
      	\caption{Modeling Pipeline}
    \label{fig:ModelPipeline}
\end{figure}

In this work pre-trained attention based large language encoders are the primary modeling component studied. Each pre-trained encoder can be easily fine-tuned by switching the output layer once weights are initialized in pre-training to one suitable for the applied task. All models are pre-trained on a Masked Language Modeling(MLM) task allowing them to capture general representations of language distributions, significantly reducing training time for researchers and increasing performance on downstream task. The modeling pipeline is shown in figure~\ref{fig:ModelPipeline}.Words are first converted to subword tokens with each pre-trained model having its own specific subword tokenizer. The two specific subword tokenization schemes applied between encoders are wordpiece tokenization and byte-pair encoding tokenization. Subword tokens are converted to embeddings and summed together with positional embeddings and encoded. For task 1 \& 2 The encoded embeddings are placed into a feed-forward network and soft-max layer for label probabilities. For task 3 support vector machine are used to get label predictions instead of a feed-forward network with a soft-max. 
All encoder implementations are imported from the the Transformers library~\cite{HUGFACE} with PyTorch as the modeling framework. Specifics about pre-training and differences for each applied encoder are shown below.

\subsubsection{Bert Base}
\label{sub:Bert}
The Bert base model applied is pre-trained on general domain corpora with all words made lowercase with fine-tuning on CMED done the same way. Bert Base contains a total of 110M parameters, made up of 12 transformer blocks each with 12 self-attention heads and hidden size for all layers in the model is 768~\cite{BERT}. The tokenization scheme applied for all Bert Base varaints(BioBert \& Bio+Clincal Bert's) follow wordpiece tokenization. Along with MLM, a next sentence prediction task is also applied as a pre-training task. The tokenization scheme applied for this encoder is wordpiece tokenization. The model was pre-trained on the following general domain corpora:
\begin{itemize}
  \item BooksCorpus - 800M Words 
  \item English Wikipedia - 2.5M Words
\end{itemize}

\subsubsection{BioBert}
\label{sub:biobert}
BioBert is the first domain-specific Bert based model pre-trained on biomedical data~\cite{BIOBERT}. The applied model is pre-trained with all lower case words so fine-tuning on CMED is done the same way. The tokenization scheme applied for this encoder is wordpiece tokenization. BioBert uses the previously explained Bert base model weights and does additionally pre-training on the following biomedical corpora:
\begin{itemize}
  \item PubMed Abstracts - 4.5B Words
  \item PMC Full-text articles - 13.5B Words
\end{itemize}

\subsubsection{Bio+Clinical Bert}
Bio+Clinical Bert~\cite{BIOCLINBERT} is another domain specific Bert model pre-trained on clinical notes. The applied model is pre-trained with all lower case words so fine-tuning on CMED is done the same way. The tokenization scheme applied for this encoder is also wordpiece tokenization. Bio+Clinical Bert uses the previously explained BioBert model weights and undergoes further pre-training on the following clinical corpus. 
\begin{itemize}
  \item MIMIC-III v1.4 database - 2M Clinical Notes
\end{itemize}
There are two variation of Bio+Clinical Bert, one trained on all MIMIC-III notes and another only on MIMIC-III discharge summaries, these will be referred to as Bio+Clinical Bert All Notes and Bio+Clinical Bert Discharge.

\subsubsection{RoBerta Base}
Robust Optimized Bert Approach known as RoBerta ~\cite{roberta} holds the same architecture of the Bert Base but undergoes modified pre-training. The modifications are as follows, training the model longer with bigger batches over more data, removing the next sentence prediction task,  training on longer sequences, and dynamically changing the masking pattern in MLM~\cite{roberta}. The tokenization scheme also differs from Bert following Byte-Pair encoding. The casesing of all words is kept in pre-training, with fine-tuning on CMED done the same way. RoBerta is pre-trained with the following corpora:
\begin{itemize}
  \item BooksCorpus - 800M Words 
  \item English Wikipedia - 2.5M Words
  \item CC-News - 60M News Articles
  \item OpenWebText - 38GB of web content
  \item Stories - 2.1B Tokens

\end{itemize}
\subsubsection{Clinical Longformer}
Clinical Longformer~\cite{ClinLong} follows the Longformer Base~\cite{Longformer} architecture with additional pre-training done on clinical text. Longformer is a Bert like architecture that allows for input sequences of up to 4069 instead of 512 limit, although in this work 512 is still applied. Longformer base takes the Roberta base checkpoint and applies a modified attention mechanism with a local windowed attention along with the global attention~\cite{Longformer}. Like RoBerta the tokenization scheme applied follows Byte-Pair encoding. Pre-training is done with keeping the case of all words, so fine-tuning on CMED is done the same way. Along with the corpora from Roberta, Clinical Longformer is pre-trained with the following corpora:
\begin{itemize}
 \item Realnews Corpora - 1.8B Tokens
  \item MIMIC-III v1.4 database - 2M Clinical Notes
\end{itemize}

\section{Experimental Results and Analysis}
\label{sec5}
\label{Experiment}
\subsection{Contexualized Medication Event Dataset}
The Contextualized Medication Event Dataset(CMED) has been released for track 1 of the 2022 n2c2 and is a dataset that captures relevant context of medication changes documented in clinical notes~\cite{N2C22022}. CMED consist of 500 clinical notes \& annotated notes with task relevant information, with 9,012 medication mentions across all the notes. The clinical notes from CMED are a portion of the data from the 2014 i2b2/UTHealth Natural Language Processing shared task~\cite{Kumer-et-al-2015}. The class distributions of CMED for all three task are shown in tables \ref{tab:CMED1Dist}, \ref{tab:CMED2Dist}, \& \ref{tab:CMED3Dist}.

\begin{table}[H] 
\caption{CMED Task 1 Distribution. Adapted from~\cite{N2C22022RES}}
\label{tab:CMED1Dist}
\centering
\begin{tabular}{llccc}
\toprule
 Label & Train & Test & Total\\
\toprule
Drug & 7229 & 1783 & 9012\\ 
\bottomrule
\end{tabular}
\end{table}

\begin{table}[H] 
\caption{CMED Task 2 Distribution. Adapted from~\cite{N2C22022RES}}
\label{tab:CMED2Dist}
\centering
\begin{tabular}{llccc}
\toprule
Label  &Train &Test & Total\\
\midrule
Disposition & 1412 & 335 & 1747\\ 
NoDisposition & 5260 & 1326 & 6586\\
Undetermined & 557 & 122 & 679\\
\bottomrule
\end{tabular}
\end{table}

\begin{table}[H] 
\caption{CMED Task 3 Distribution. Adapted from~\cite{N2C22022RES}}
\label{tab:CMED3Dist}
\centering
\begin{tabular}{llccc}
\toprule
Context Dimension&Label  &Train &Test & Total\\
\midrule
\textbf{Action} & Start & 568 & 131 & 699\\ 
\textbf{} & Stop & 340 & 67 & 407\\ 
\textbf{} & Increase & 129 & 22 & 151\\ 
\textbf{} & Decrease & 54 & 13 & 67\\ 
\textbf{} & Unique Dose & 285 & 88 & 373\\ 
\textbf{} & Other Change & 1 & 0 & 1\\ 
\textbf{} & Unknown & 35 & 14 & 49\\
\midrule
\textbf{Temporality} & Past & 744 & 173 & 917\\
\textbf{} & Present & 494 & 132 & 626\\ 
\textbf{} & Future & 145 & 29 & 174\\ 
\textbf{} & Unknown & 29 & 1 & 30\\ 
\midrule
\textbf{Actor} & Physician & 1278 & 311 & 1589\\
\textbf{} & Patient & 106 & 17 & 123\\ 
\textbf{} & Unknown & 28 & 7 & 35\\ 
\midrule
\textbf{Certainty} & Certain & 1176 & 281 & 1457\\
\textbf{} & Hypothetical & 134 & 33 & 167\\ 
\textbf{} & Conditional & 100 & 15 & 115\\ 
\textbf{} & Unknown & 2 & 6 & 8\\ 
\midrule
\textbf{Negation} & Negated & 32 & 6 & 38\\
\textbf{} & Not Negated & 1380 & 329 & 1709\\ 
\bottomrule
\end{tabular}
\end{table}

\subsection{Processing}
\label{sec:proc_methods}
\begin{figure}[H]
	 \centering
    	 \includegraphics[width=1.1\linewidth]{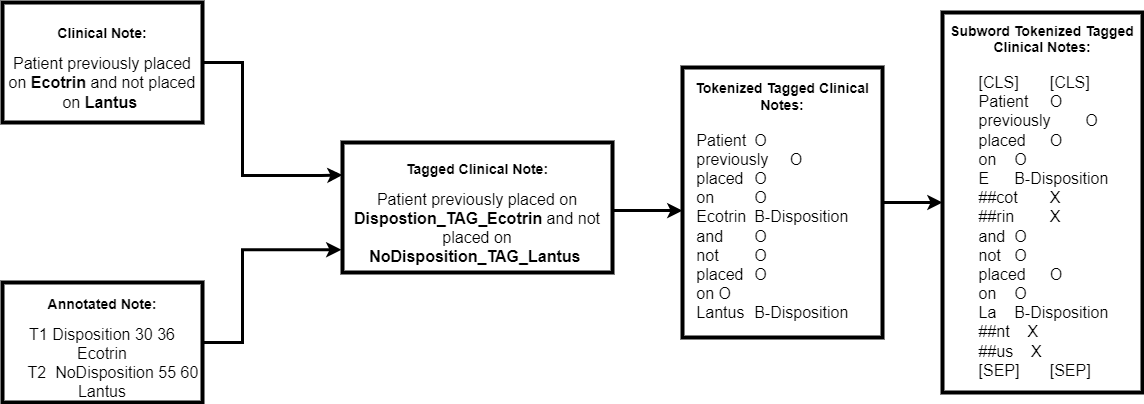}
      	\caption{Pre-Processing Pipeline}
    \label{fig:PreProcPipeline}
\end{figure}

 The pre-processing pipeline is shown in figure~\ref{fig:PreProcPipeline} and in the following order consist of annotation parsing, EHR tagging, EHR sentence/section segmentation, word tokenization, and subword tokenization. All pre-processing is done with Python and is largely rule-based with the exception of sentence segmentation and word tokenization being done using pre-trained models from the PunkT~\cite{PUNKT} package and wordpiece tokenization or byte-pair tokenization done with pre-trained models paired with each applied encoder. For task 1 and task 2 the data is pre-processed into a BIO tag format for Named Entity Recognition. For task 3 a BIO format is not used, since only medications that have a disposition label from task 2 are used for training data, these medications are given just their class labels for each context dimension.  

\begin{figure}[H]
	 \centering
    	 \includegraphics[width=1\linewidth]{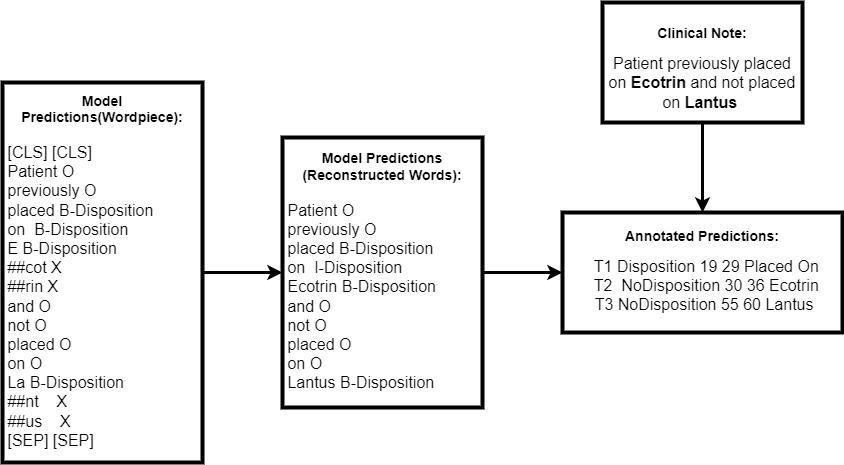}
      	\caption{Post Processing Pipeline}
    \label{fig:PostProc}
\end{figure}

The post-processing pipeline is shown in figure~\ref{fig:PostProc} and consist of word reconstruction from subwords and placing predictions in an annotated format for compatibility with an evaluation script released for the 2022 N2C2 competition.

For training in task 1 and task 2 the entire model with encoder and output layers are fine-tuned with CMED with all words from the training portion used. For task 3 only the SVM's are trained using the embeddings from each pre-trained encoder, with an SVM for each context dimension.
\subsection{Evaluation Metrics}
\label{sec:metrics}
The evaluation metrics used in this study are precision, recall and F1-Score. Precision is defined as the probability that an object is relevant given that it is returned by the system~\cite{METRICS} or as the number of true positives(correct labels) for a particular label out of all entities returned by a system and is shown in equation~\ref{equ:precision}.

\begin{equation}
\label{equ:precision}
	Precision = \frac{TP}{TP+FP}.
\end{equation}
Recall is defined as the probability that a relevant object is returned by a system~\cite{METRICS} or the number of true positives(correct labels) for a particular label out of all the entities that actually have that been returned by a system and is shown in equation~\ref{equ:recall}.
\begin{equation}
\label{equ:recall}
	Recall = \frac{TP}{TP+FN}.
\end{equation}
An F-Score is a measure of a models accuracy on a dataset and is defined as the harmonic mean between precision and recall and is shown in equation~\ref{equ:MicroF1},
\begin{equation}
\label{equ:MicroF1}
	FScore = \frac{2 \times Precision \times Recall}{Precision + Recall}.
\end{equation}

Predictions are based on character positions in EHR's and strict and lenient matching are considered for calculating each evaluation metrics, where the strict matching refers to if an prediction matches the exact correct character position associated with the medication, while with lenient matching if there is overlap in the prediction and correct character positions, then the prediction counts as correct.

\subsection{Results and Analysis}

\subsubsection{Task 1: Medication Detection}
Results for task 1 are shown in table~\ref{Task1Res}. On average models achieve a strict F-Score of 0.9226 and lenient F-Score of 0.9581. Bio+Clinical Bert Discharge shows the best performance for this task with a strict score 0.0129 higher than the average at 0.9355 and lenient score .0089 higher than the average at 0.9669. RoBerta base shows the worst performance for this task with a a strict F-Score of 0.9075, which is .0151 less than the average  and lenient F-Score of 0.9476 which is 0.0105 less than the average. Across all metrics models pre-trained on clinical data(Bio+Clinical Bert's and Clinical Longformer) show the best performance, while models trained on only general domain data achieve the worst performance(Bert Base and RoBerta Base).

\begin{table*}[!ht]
\caption{Performance comparison on medication detection}
\label{Task1Res}
\centering
\begin{tabular}{|l|ccc|ccc|}
\hline
\textbf{Model}&\multicolumn{3}{c|}{\textbf{Strict Evaluation Performance}}&\multicolumn{3}{c|}{\textbf{Lenient Evaluation Performance}} \\
\hline
\textbf{Pretrained Model}  & \textbf{Precision} & \textbf{Recall} & \textbf{F-Score} & \textbf{Precison} & \textbf{Recall} & \textbf{F-Score} \\
\hline
$BERT base$	& 0.8939 & 0.8365 & 0.9147 & 0.9297 & 0.9739 & 0.9513\\
$BioBERT$	& 0.9154 & 0.9235 & 0.9239 & 0.9521 & 0.9700 & 0.9610 \\
$Bio+Clinical Bert All Notes$&\textbf{0.9281} & 0.9291 & 0.9255 & \textbf{0.9612} & 0.9688 & 0.9650\\
$Bio+ClinicalBert Discharge$&0.9252 & 0.9461 &\textbf{0.9355} &0.9562  &\textbf{0.9779} &\textbf{0.9669} \\
$RoBerta$	& 0.8925 & 0.9229 & 0.9075 &0.9320  &0.9637 &0.9476 \\
$Clinical Longformer$ & 0.9097	& \textbf{0.9478} & 0.9284  &0.9374 &0.9768  & 0.9567 \\
\hline
\end{tabular}
\end{table*}

\subsubsection{Task 2: Medication Event Classification}
Results for task 2 are shown in table~\ref{Task2Res}. On average models achieve a strict F-Score of 0.8280 and lenient F-Score of 0.8614. Clinical Longformer is the top performing model for this task with a strict F-Score of 0.8515, which is 0.0235 higher than the average and lenient F-Score 0.8793 which is 0.0179 higher than the average. Like task 1, the model pre-trained with clinical data are the top performers across all metrics with specifically Clinical Longformer and Bio+Clinical Bert Discharge having the top scores for all metrics. Also like task 1 models pre-trained on general domain data only achieve the lowest performance on this task.

\begin{table*}[!ht]
\caption{Performance comparison on medication event classification}
\label{Task2Res}
\centering
\begin{tabular}{|l|ccc|ccc|}
\hline
\textbf{Model}&\multicolumn{3}{c|}{\textbf{Strict Evaluation Performance}}&\multicolumn{3}{c|}{\textbf{Lenient Evaluation Performance}} \\
\hline
\textbf{Pretrained Model}  & \textbf{Precision} & \textbf{Recall} & \textbf{F-Score} & \textbf{Precision} & \textbf{Recall} & \textbf{F-Score} \\
\hline
$BERT base$	&0.7922	&0.8295&0.8104&	0.8268&	0.8657&	0.8458\\
$BioBERT$	&0.8125& 0.8272&0.8198 & 0.8464	&0.8618&0.8540\\
$Bio+Clinical Bert All Notes$&0.8268&	0.8329&	0.8298 & 0.8650	&0.8714&0.8682\\
$Bio+ClinicalBert Discharge$&\textbf{0.8370}	&0.8555	&0.8462 & \textbf{0.8659} &0.8850&0.8753\\
$RoBerta$	& 0.7971& 0.8283  &0.8103 &0.8322& 0.8601&0.8459\\
$Clinical Longformer$ & 0.8346	& \textbf{0.8691} & \textbf{0.8515}  &0.8618 &\textbf{0.8975}  & \textbf{0.8793} \\
\hline
\end{tabular}
\end{table*}

\subsubsection{Task 3: Multi-dimensional Context Classification}
Results for task 3 are shown in table~\ref{Task3Res}, Strict and lenient scores are not considered for this task as position of medications are not being predicted for this task. Results are also taken from two perspectives, the first like the other task is the overall F-score across context dimensions, the second perspective considers all context dimensions predictions combined and counts prediction to be correct if all dimensions have correct labels. For the overall performance on average models achieve and F-Score of 0.5986 and for the combined performance models achieve an average score of 0.1754. Bert Base shows the best performance across all categories largely outperforming other models with an overall score of 0.7387, which is 0.1401 higher than the average and a combined F-Score of 0.3006, which is 0.1251 higher than the average. RoBerta and Clinical Longformer show the poorest performance, largely under-performing all Bert Base variants.  

\begin{table*}[!ht]
\caption{Performance comparison on multi-dimensional context classification}
\label{Task3Res}
\centering
\begin{tabular}{|l|ccc|ccc|}
\hline
\textbf{Model}&\multicolumn{3}{c|}{\textbf{Overall  Performance}}&\multicolumn{3}{c|}{\textbf{Combined Performance}} \\
\hline
\textbf{Pretrained Model}  & \textbf{Precision} & \textbf{Recall} & \textbf{F-Score} & \textbf{Precision} & \textbf{Recall} & \textbf{F-Score} \\
\hline
$Bert base$	&\textbf{0.7569}&\textbf{0.7188}&\textbf{0.7387}	&\textbf{0.3091}&\textbf{0.2925}&\textbf{0.3006}\\
$BioBert$& 0.7104&0.6722&0.6908&0.2366& 0.2239&0.2301 \\
$Bio+Clinical Bert All Notes$&0.6826	&0.6460&0.6638& 0.2303	&0.2179&0.2239\\
$Bio+ClinicalBert Discharge$	&0.6650&0.6293&0.6466 & 0.2114&0.2000&0.2055\\
$RoBerta$ &0.4038 & 0.3821&0.3926&0.0410 &0.0388&0.0399 \\
$Clinical Longformer$& 0.4719&0.4466 &0.4589&0.0536&0.0507&0.0521 \\
\hline
\end{tabular}
\end{table*}

\section{Related Works}
\label{sec3}
\label{Related}

Since the release of the transformer~\cite{TRANSFORMER} model in 2017 a number of studies have been done around their application on clinical text. The GPT variant GPT2 has been applied for generating synthetic EHR data as a solution to overcoming privacy concerns with EHR's~\cite{Libbi-et-al-2021}. Bert has been popular in information extraction, where like in this work the encoder is fine-tuned on an EHR dataset. This strategy has been applied in task such as predicting depression~\cite{BERTDepr}, identifying events where patients have injuries due to falls~\cite{BERTFall}, and identifying adverse drug events~\cite{BertClinCon}. Along with fine-tuning transformer models for task, several models like Clinical Longformer and Bio+Clinical Bert have been pre-trained with masked langauge modeling to create domain specific models. Some of these pre-trained models include MedBert~\cite{MEDBERT}, CancerBert~\cite{cancerbert}, \& BlueBert~\cite{bluebert}.

Before transformer based models Recuurent Neural Network models were the standard approach for modeling EHR's. Track 2 of the 2018 N2C2 like track 1 of the 2022 N2C2 involved medication related information extraction and the RNN variant Bidirectional long short-term memory with Condition Random Fields(BiLSTM CRF's), was a popular modeling choice, with 9 of the top teams incorporating them in their system.~\cite{Henry2020}. BiLSTM-CRFs use a BiLSTM to create a series of state representations that are then used as input into a CRF for labeling~\cite{Henry2020}. Static embedding methods had been applied to accompany the BiLSTM-CRF's ,one popular method was to use the MIMIC-III dataset to create pre-trained word embeddings with the Word2Vec package another was to use GloVe embeddings\cite{Dai2020}.


\section{Conclusion}
\label{sec7}

\label{Conclusion}
In this work a comparative analysis is presented amongst attention based large language models for the following EHR information extraction task: medication detection, medication event classicatition, and  multi-dimensional medication event context classification. The applied models are Bert Base, BioBert, 2 variations of Bio+Clinical Bert, RoBerta Base, and Clinical LongFormer.  The models are evaluated on all task with recall, precision, and F-Score as metrics. For medication detection models pre-trained on clincial data show the best performance with Bio+Clinical Bert pre-trained on MIMIC III Discharge notes achieveing the highest performance with a strict F-Score of 0.9355 and lenient F-Score of 0.9669. For medication event classication, like medication detection, models pre-trained with clinical data achieve the best performance with the highest F-Scores coming from Clinical LongFormer with a strict F-Score of 0.8515 and lenient F-Score of 0.8793. For multi-dimensional medication event context classification, Bert base achieves the best performance. For this task two perspectives are considered, the first is the overall F-Score, for this Bert Base achieves a score of 0.7387, the second perspective is scoring on how many predictions have all context dimension prediction correct for a medication, and for this Bert Base achieves a score of 0.3091. For future work, increasing performance on the third task is considered, especially for generating correct predictions across all dimensions. Methods around data augmentation are of interest, there is a lack of data for this task, with some classes having as little as 1 label in the training set. 

\section*{Acknowledgment}
\label{acknowledgement}
{This research work is supported in part by the U.S. Office of the Under Secretary of Defense for Research and Engineering (OUSD(R\&E)) under agreement number FA8750-15-2-0119. The U.S. Government is authorized to reproduce and distribute reprints for governmental purposes notwithstanding any copyright notation thereon. The views and conclusions contained herein are those of the authors and should not be interpreted as necessarily representing the official policies or endorsements, either expressed or implied, of the Office of the Under Secretary of Defense for Research and Engineering (OUSD(R\&E)) or the U.S. Government.}

\ifCLASSOPTIONcaptionsoff
  \newpage
\fi


\bibliographystyle{ieeetr}
\bibliography{references}
\end{document}